\documentclass[9pt,technote,compsoc]{IEEEtran}
\usepackage{amssymb}  

\ifCLASSOPTIONcompsoc
  \usepackage[noadjust,nocompress]{cite}
\else
  \usepackage{cite}
\fi
\ifCLASSINFOpdf
  \usepackage[pdftex]{graphicx}
  
  \graphicspath{figs/}
  \DeclareGraphicsExtensions{.pdf,.jpeg,.png}
\else
\fi
\usepackage[cmex10]{amsmath}
\ifCLASSOPTIONcompsoc
  \usepackage[caption=false,font=normalsize,labelfont=sf,textfont=sf]{subfig}
\else
  \usepackage[caption=false,font=footnotesize]{subfig}
\fi
\begin{document}
%
\title{Learning Invariant Color Features for\\ Person Re-Identification}
%
%
%
%

\author{Rahul~Rama~Varior, ~\IEEEmembership{Student Member,~IEEE,}
        ~Gang~Wang,~\IEEEmembership{Member,~IEEE} ~Jiwen~Lu,~\IEEEmembership{Member,~IEEE}
\IEEEcompsocitemizethanks{\IEEEcompsocthanksitem R. Rama Varior and G. Wang are with the School of Electrical and Electronic Engineering, Nanyang Technological University, Singapore, 639798.\protect
E-mail: \{rahul004,wanggang\}@ntu.edu.sg.\protect }
\IEEEcompsocitemizethanks{\IEEEcompsocthanksitem G. Wang and J. Lu are with the Advanced Digital Sciences Center, Singapore, 138632. \protect
Email: \{gang.wang, jiwen.lu\}@adsc.com.sg.\protect

}
\thanks{}}

\IEEEtitleabstractindextext
{%
\begin{abstract}
\makebox[212pt][s]{Matching people across multiple camera views known as } \par
\makebox[252.5pt][s]{person re-identification, is a challenging problem due to the change in} \par
\makebox[252.5pt][s]{visual appearance caused by varying lighting conditions. The perceived} \par
\makebox[252.5pt][s]{color of the subject appears to be different with respect to illumination.} \par
\makebox[252.5pt][s]{Previous works use color as it is or address these challenges by} \par
\makebox[252.5pt][s]{designing color spaces focusing on a specific cue. In this paper, we} \par
\makebox[252.5pt][s]{propose a data driven approach for learning color patterns from pixels} \par
\makebox[252.5pt][s]{sampled from images across two camera views. The intuition behind} \par
\makebox[252.5pt][s]{this work is that, even though pixel values of same color would be} \par
\makebox[252.5pt][s]{different across views, they should be encoded with the same values.} \par
\makebox[252.5pt][s]{We model color feature generation as a learning problem by jointly} \par

\makebox[252.5pt][s]{learning a linear transformation and a dictionary to encode pixel values.} \par
\makebox[252.5pt][s]{We also analyze different photometric invariant color spaces.} \par
\makebox[252.5pt][s]{Using color as the only cue, we compare our approach with all the photo-} \par
\makebox[252.5pt][s]{metric invariant color spaces and show superior performance over all} \par
\makebox[252.5pt][s]{of them. Combining with other learned low-level and high-level features,} \par

\makebox[252.5pt][s]{we obtain promising results in ViPER, Person Re-ID 2011 and CAVIAR4-} \par
\makebox[25.25pt][s]{REID datasets.} \par

\end{abstract}

}

\maketitle

\IEEEdisplaynontitleabstractindextext

%
\IEEEpeerreviewmaketitle

\vspace*{-0.65cm}
\section{Introduction}
\label{sec:intro}
\begin{figure}
\centering
\includegraphics[width=0.1\linewidth]{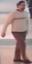}
\includegraphics[width=0.1\linewidth]{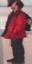}
\includegraphics[width=0.075\linewidth]{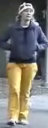}
\includegraphics[width=0.075\linewidth]{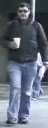}
\includegraphics[width=0.1\linewidth]{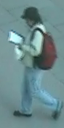}
\includegraphics[width=0.1\linewidth]{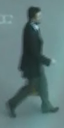} \\
\vspace{3pt}
\includegraphics[width=0.1\linewidth]{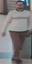}
\includegraphics[width=0.1\linewidth]{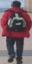}
\includegraphics[width=0.075\linewidth]{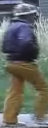}
\includegraphics[width=0.075\linewidth]{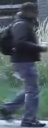}
\includegraphics[width=0.1\linewidth]{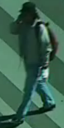}
\includegraphics[width=0.1\linewidth]{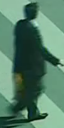}
\caption{Some examples from CAVIAR4REID, ViPER and Person Re-ID 2011 datasets. Images clearly show the appearance changes due to the environmental variables - Illumination, Shading, Camera view angle. {\bf Best viewed in color}}
\label{samplefigs}
\end{figure}

\IEEEPARstart{M}{atching} pedestrians across multiple CCTV cameras have gained a lot of interest in recent years. Despite several attempts\cite{kviatkovsky2013color,custompict,bicovma2012} to address these challenges, it largely remains challenging mainly due to the following reasons. First, the images are captured under different lighting conditions. Therefore the perceived color of the subject appears to be different with respect to the illumination. Second, from surveillance cameras, no biometric aspects are available\cite{custompict}. Third, most often, the surveillance cameras will be of lower resolution\cite{kviatkovsky2013color}. Figure \ref{samplefigs} shows some examples of images from different datasets.

Modern person re-identification systems primarily focus on two aspects. (1) A feature representation for the probe and gallery images and (2) a distance metric to rank the potential matches based on their relevance. In the first category, majority of the works has been done on designing low level features. 
Since each of the features capture different aspects of the images, usually a combination of these features are used to obtain a richer signature. In the second category, the person re-identification is formulated as a ranking 
or a metric learning 
problem. 
In this work, we focus on the feature representation aspect, specifically on color based features.

Color based features have been proven to be an important cue for person re-identification\cite{kviatkovsky2013color}. An interesting insight on the importance of color features was demonstrated in an experiment conducted by Gray and Tao\cite{viewpointinvariant}. They used AdaBoost for giving weights to the most discriminative features. They observed that, over $75$ percent of the classifier weights were given to color based features. These observations support the fact that color has to be given much more attention than other handcrafted features based on shape, texture and regions. But due to the illumination variations across the camera views, the perceived color of same parts for a particular person appear to be different.
Taking this observation into consideration and as validated from our experiments, we suggest that using color features {\it as it is}, i.e. RGB, HSV or YUV color histogram representation will not be adequate to achieve an illumination invariant representation. Hence we propose a data driven multilayer framework that learns invariant color features from raw pixel values as opposed to histogram or other color based handcrafted features.

The proposed framework aims at learning an invariant representation for images from both camera views by transforming all the pixels to a color-constant space. In the color-constant space, the pixels are invariant to the camera and other environmental variables. Instead of modeling each of the variables, we exploit a data driven framework to explore the structures and patterns inherent in the image pixels. Feature learning approaches come in handy in this situation. We use an auto-encoder based framework to transform the 3-dimensional RGB pixel values to a higher dimensional space first and encode them using a dictionary. These encoded values are pooled over a region and concatenated to form the final representation of an image. This framework can be extended to a multilayer structure for learning complementary features at a higher level. 

Experiments were conducted on publicly available datasets such as ViPER\cite{viper}, Person Re-ID 2011\cite{hirzer11a} and CAVIAR4REID\cite{custompict}. From the results, it can be inferred that (1) by learning illumination invariant color features, significant improvement can be achieved in the results when compared with the traditional color histograms and other handcrafted features; and (2) when combined with other types of learned low-level and high-level features, it can achieve promising results in several benchmark datasets.

The rest of this paper is organized as follows. Section 2 reviews some of the related works in color constancy, person re-identification and feature learning. Section 3 describes the motivation and the major contributions of this work. Section 4 describes the framework for learning the invariant color features. In section 5, we  demonstrate the experimental evaluation of our method and compare with the other competing methods for person re-identification. In section 6, we perform an analysis of the obtained results and Section 7 concludes this paper. 

\section{Related Works}

\subsection{Color Constancy}
\label{subsec:cc}
Human perceptual system has the ability to ensure that the perceived color of an object remains relatively constant even under varying illumination\cite{fastimplofccmorel2009}.  Land and McCann proposed the Retinex theory\cite{retinexland71} to explain this perceptual effect. 
In one of the pioneering works\cite{forsythcolorconst}, Forsyth proposed the CRULE and MWEXT algorithms to achieve color constancy in Mondriaan world images by estimating the illuminant based on the information obtained from images such as reflectances and possible light sources. 
For a detailed overview of the color-constancy algorithms derived from the Retinex theory and \cite{forsythcolorconst}, we refer the reader to \cite{ccsurvey2011,ebner2007color,hordley2006scene}.

All of the aforementioned works and the derived works are based on specific assumptions since the color constancy is an under-constrained problem
\cite{forsythcolorconst, hordley2006scene}. 
For example, in \cite{forsythcolorconst}, the main assumption is constrained gamuts, i.e., the limited number of image colors which can be observed under a specific illuminant. Several other assumptions were based on the  distribution of colors that are present in an image (e.g., white patch, gray-world and gray-edge). Majority of these works vary in their assumptions and therefore no color constancy algorithm can be considered as universal. 

Several works were done focusing on color features for object recognition. In one of the earliest works, Swain and Ballard\cite{swainballardcolorhistogram} identified that color histograms were stable representations over change in views. 
Gevers and Smeulders\cite{Geverscolorbasedobj} analysed different color spaces to achieve invariance to a substantial change in viewpoint, object geometry and illumination. But it was observed that the object recognition accuracy degrades substantially for all of the color spaces with a change in illumination color. 

\begin{figure}[!t]
\centering
\subfloat[]{\includegraphics[trim = 28.57mm 112.83mm 114.73mm 106.36mm, clip,width=0.4\linewidth]{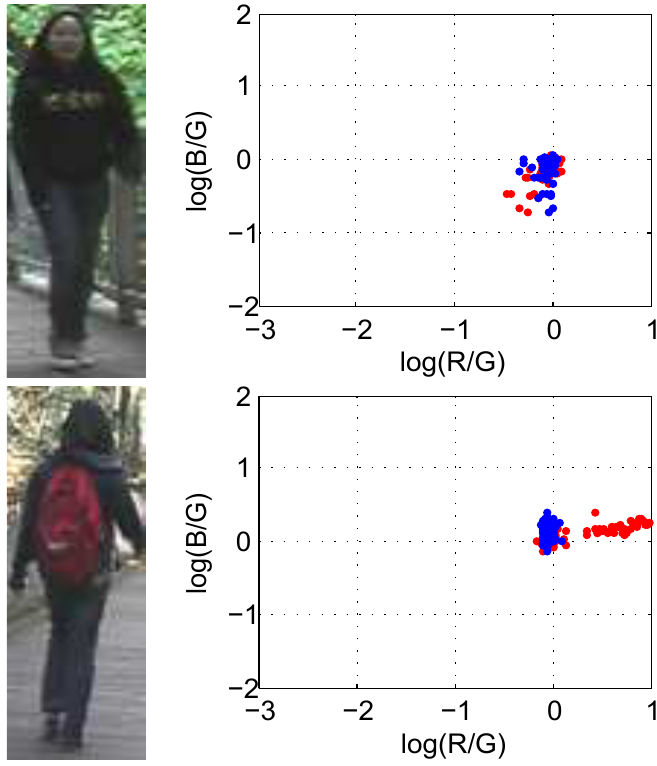}
\label{fig_first_case}}
\hfil
\hspace{1cm}
\subfloat[]{\includegraphics[trim = 114.3mm 112.83mm 27.432mm 106.36mm, clip,width=0.4\linewidth]{final.pdf}
\label{fig_second_case}}
\caption{Two examples to illustrate the difference in color clouds for the same person in different views due to appearance change in the second view. In (a), due to the presence of additional object(bag), the shape of the color clouds vary significantly. In (b), the color of the upper body has changed across the views which result in a different shape for the corresponding color clouds as shown in the plot. {\bf Best viewed in color}}
\label{fig_colorclouds}
\end{figure}

A Diagonal Matrix Transform (DMT) is the basis of majority of the 
works\cite{berwicklee,healey1994global,Geverscolorbasedobj,kviatkovsky2013color} on color constancy. 
To improve the performance of DMT, spectral sharpening\cite{Finlayson94} derived for each camera can be incorporated. 
Berwick and Lee\cite{berwicklee} proposed a log-chromaticity color space to achieve specularity, illumination color and illumination pose invariance. A recent work\cite{kviatkovsky2013color} make use of the log-chromaticity color space to achieve invariant color features for person re-identification. The assumption in \cite{kviatkovsky2013color} is that the shape of the color cloud is sufficiently preserved in the log$\frac{R}{G}$, log$\frac{B}{G}$ space. But this assumption is violated in many real world images. The color cloud is formed based on sampled observations from upper body and lower part of the body.  \figurename{\ref{fig_colorclouds}} shows some examples of those observations from the ViPER dataset. It can be seen that in a different view, the upper part of the body may have different colors for the same subject. In this paper, data driven techniques are used to discover a color constant space in contrast to methodologies based on strong assumptions.
\subsection{Person Re-Identification}

Person re-identification research has received a good amount of attention in recent years. As mentioned in section \ref{sec:intro}, existing works focus on the different steps that need to be taken for dealing with this problem. Majority of the works\cite{custompict,sdalf,bicovma2012,viewpointinvariant,zhao2013unsupervised} predominantly focus on the first step, i.e. designing features based on texture, color, shape, regions and interest points. Since the primary focus of this work is on color based feature design, a complete evaluation of all of the aforementioned features is beyond the scope of this paper. To obtain the global chromatic content, most of the works uses the color histogram features in RGB, HSV or YUV space. These color spaces do not possess the property of illumination invariance. In addition to a weighted HSV histogram, Maximally Stable Color Regions(MSCR) are also used in \cite{sdalf} to obtain the per-region color displacement. 

In a relatively closer work, Porikli\cite{btfporikli2003} and Javed\cite{btfjaved2005} proposed a Brightness Transfer Function(BTF) to find a transformation that maps the appearance of an object in one camera view to the other. But it should be noted that, the system has to be re-trained each time the illumination changes. In addition to that, the method adopts normalized histograms of object brightness values for BTF computation. Therefore, a pixel level correspondence cannot be achieved. 

It is important to note that all the aforementioned features for person re-identification are handcrafted focusing on specific cues. However, in this work we propose a model to learn color based features for a pair of camera to achieve the invariance properties across the two different views based on the intuition that the features should be invariant to change in environment and camera variables. To the best of our knowledge, this is the first work that focuses on a data driven learning of low-level color features. Since the framework is based on learning from the data, the scope of this work is beyond person re-identification. 



\subsection{Feature Learning}

Recent researches have shown a growing interest in unsupervised feature learning methods such as auto-encoders\cite{vincent2008extracting}, sparse coding\cite{yang2009linear} and Deep Belief Nets\cite{hinton2006fast} since they are data driven and can be generalized to a larger extent. Since no handcrafted feature can be considered as universal, learning relations from data can be advantageous. 

Modeling complex distributions and functions have been a bottleneck in machine learning. Recent studies in deep learning indicate that such deep architectures can efficiently handle these challenges and have shown that better generalization can be obtained. Several successful algorithms have been proposed\cite{hinton2006fast},\cite{hinton2006reducing,lee2008sparse,bengio2007greedy,bengio2009learning} to train large networks such as deep belief networks and stacked auto-encoders. The intuition behind using large networks is that, to learn a complex function that computes the output from input, automatically learning features at multiple layers of abstraction can help to a large extend\cite{bengio2009learning}. Additionally, biological evidences substantiate that in the visual cortex, recognition happens at multiple layers\cite{serre2007quantitative}. However, all the above works focus on learning edges at different orientations in the first layer and higher level patterns in the further layers. In contrast to this, the proposed work addresses the problem of learning color features from data. We also propose a new method to learn a transformation and encoding simultaneously at a pixel level to achieve color constancy.


\label{subsec:fl}
\section{Motivation}
\label{sec:motiv}

Even though the importance of color features is proven in \cite{viewpointinvariant}, color features were not given much attention in person re-identification. Existing works use color {\it as it is}, i.e. computing the histogram in designed color spaces without any processing. However, varying lighting conditions affect the robustness of such designed features. These variations can be considered as noises that corrupt the actual pixel values. While designing color features, the performance will be affected unless these noises are handled well. The existing methods for achieving color constancy are based on strong assumptions about the statistics of color distribution, surfaces and its reflectance properties. Hence, a histogram representation in such a {\it weakly} corrected space will not be robust enough. 

Hand-engineered features focus on particular cues and adds more complexity to the system. Modeling each of the camera parameters and other environmental variables explicitly is practically impossible. Our intuition is that a {\it data driven} feature learning approach can discover a good intermediate representation from the input pixels. This means that, the relationship between the images across views can be learned by sampling observations from the images themselves. A robust representation should capture a certain amount of {\it information}, i.e. stable structures and patterns from the observed input. Though the inputs are corrupted, it can be reasonably assumed that there exists a space where the color patterns are invariant to these variables. A manifold learning interpretation of this problem is given in \cite{vincent2008extracting}. The corrupted inputs will lie away from the manifold and the objective is to learn a transformation to project them back to the manifold, in this case, the color constant space. With this assumption, we use a linear auto-encoder to discover such patterns from the data with the objective that in the transformed space, the representation for pixels of same colors should be as close as possible.

Many researchers have empirically found that an encoding schemes for quantizing each local descriptors is essential for good performance. 
Sparse coding \cite{yang2009linear, yu2009nonlinear} has been found to achieve state-of-the-art performance for many classification problems. In this work, a sparse coding technique is adopted to encode each pixel by a set of dictionaries or codebooks. 
Previous works consider feature computation and encoding independently. However, these two approaches should be consistent with each other. Therefore, we propose a joint learning framework to obtain a transformation and the codebook simultaneously while enforcing the final encoded representation of each pixel belonging to the same color to be same. As validated from our experiments, we observe that the joint learning framework helps to obtain a robust representation and boost the performance significantly. 


Considering all the aforementioned challenges, we develop a data driven joint learning framework to handle these variables effectively. In summary, the contributions of our work are as follows

\begin{figure*}[!t]
\centering
\subfloat[View A (Image and the Extracted Patch)]{\includegraphics[trim = 5mm 244mm 146mm 0mm, clip,width=0.2\linewidth]{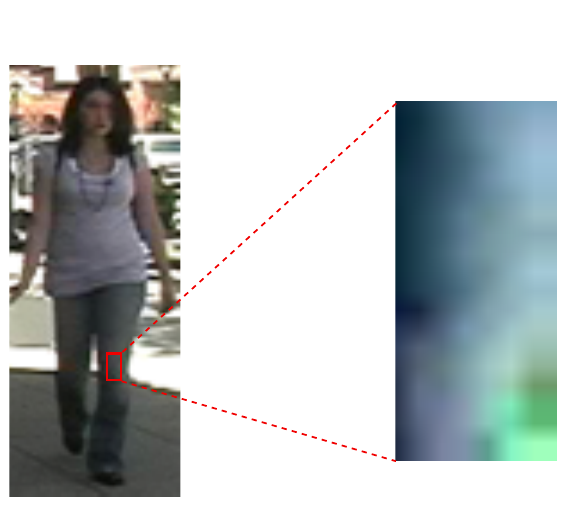}
\label{fig_imga}}
\hfil
\subfloat[RGB Histogram]{\includegraphics[trim = 62.5mm 244mm 78mm 0mm, clip,width=0.2\linewidth]{sparsecodevshist.pdf}
\label{fig_imgb}}
\hfil
\subfloat[Encoding using our approach]{\includegraphics[trim = 134mm 240mm 3mm 0mm, clip,width=0.2\linewidth]{sparsecodevshist.pdf}
\label{fig_imgc}} \\

\vspace{-0.8cm}
\subfloat[View B (Image and the Extracted Patch)]{\includegraphics[trim = 5mm 187mm 146mm 57mm, clip,width=0.2\linewidth]{sparsecodevshist.pdf}
\label{fig_imgd}}
\hfil
\subfloat[RGB Histogram]{\includegraphics[trim = 62.5mm 187mm 78mm 57mm, clip,width=0.2\linewidth]{sparsecodevshist.pdf}
\label{fig_imge}}
\hfil
\subfloat[Encoding using our approach]{\includegraphics[trim = 134mm 183mm 3mm 57mm, clip,width=0.2\linewidth]{sparsecodevshist.pdf}
\label{fig_imgf}}
\caption{(a) Training image from view A and a patch sampled from that image. (b) RGB histogram (216 dimensional) of the extracted patch. (c) Randomly sampled pixel encoded using our approach. (d) Training image from view B and a patch sampled from the corresponding region as patch from Image in (a). (e) RGB histogram (216 dimensional) of the extracted patch. (f) Encoding of a randomly sampled pixel from this patch. It can be seen from (b) and (e), the RGB histogram representation for the corresponding sampled patches are not close to each other. (c) and (f) shows the result of the encoding using our approach. It can be seen that, using our formulation, the obtained encoding is very close to each other for corresponding pixels in the sampled patch. We learn the transformation and Dictionary jointly so that the encoding is same for the two sampled patches. {\bf Best viewed in color}}
\label{fig_sparsecodevshist}
\end{figure*}

\begin{itemize}
  \item We propose a novel data driven approach for learning inter camera illumination invariant color features from the pixels sampled from matching pair of images. In contrast with the previous works such as color histograms, normalized color spaces and BTF, our approach is more robust and represents the color features in an efficient and discriminative way.
%
  \item We propose a joint learning framework which solves the coupled problem of learning a linear auto-encoder transformation and a dictionary to encode the features. Previous works make use of an independent strategy to obtain features and encode them.

  \item We show that color as a single cue can bring a good performance that beats several hand-engineered features designed for person re-identification and when combined with other types of learned low-level and high-level features, it can achieve promising performance in several challenging datasets.
\end{itemize}

%
%
%
%
%
%
%
%
%
%
%
%

\section{Approach}

As mentioned in the previous sections, the appearance of color changes across camera views due to a stark change in illumination. \figurename{\ref{fig_imga}} and \figurename{\ref{fig_imgd}} show an example of such changes in the appearance. It can be seen that the patches sampled, as shown in \figurename{\ref{fig_imga}} and \figurename{\ref{fig_imgd}}, appears to be of different color. Histogram of the sampled patches appears to be as shown in the \figurename{\ref{fig_imgb}} and \figurename{\ref{fig_imge}} respectively. Finlayson et al.\cite{Finlayson94} have shown that the diagonal model is an accurate model to achieve color-constancy for narrow-band(sensitive to single wavelength) imaging sensors. Practically, such sensors do not exist and the diagonal model is thus considered as an approximate model to correct the images for its illumination changes. 

Patches sampled from the images are pre-processed using L2-norm as mentioned in \cite{weakinvariancedrew1998}. We use a linear auto-encoder to transform the pixels into a rich higher dimensional space. These transformed pixel values are encoded using a sparse coding technique to attain more discriminative information. The objective behind the encoding is that the encoded values for corresponding pixels should be the same(or very close). To achieve consistency between the linear transformation and the encoding, we adopt a joint learning strategy to optimize the linear auto-encoder transformation and dictionary learning simultaneously. The parameters are updated alternatively to find the optimal mapping and encoding for the pixel values. More details are given in the subsequent sections.


\subsection{Training Patch Collection}

To train the system, patches were extracted from the training images manually. As a selection principle, we carefully choose patches so that sampled patches are distributed among different colors under varying illuminations. Background information were discarded and patches were cropped from corresponding parts of the same subject in two views where the color appears to be different. This is very important to create an optimal model and it can be seen from our experiments that this model based on these sampled patches gives a very good performance. From each of the training image pairs in each dataset, we sample 3 pairs of patches by following the criteria mentioned above. \figurename{\ref{fig_imga}} and \figurename{\ref{fig_imgd}} show an example of pair of patches sampled from the ViPER dataset.



\subsection{Objective Formulation and Optimization}

Intuitively, in a color constant space, the pixel values of same color should be same for both the images. Therefore, the objective of the proposed framework is to compute a representation so that the pixel values are close to the corresponding pixels in the matching image. This objective is formulated by making use of an auto-encoder and sparse encoding technique by enforcing the final encoded pixels of same color to have same values. As mentioned in section \ref{sec:motiv}, auto-encoder captures the stable structures and patterns in the data and projects it into a  color constant space. At the same time, taking the performance into consideration, a sparse encoding is also applied to the descriptors to represent them compactly. Mathematically: 


\begin{equation}
\label{eqn_test}
\mathop\text{minimize }_{\bf W_1, W_2, D, \alpha_1, \alpha_2} {\ell_{ae} + \ell_{sc} + \varepsilon_{en} + \Omega}, \\
\end{equation}
\begin{eqnarray}
\ell_{ae} &=& \frac{1}{m} \left \| \left ( W_2'\left ( W_1' X + b_1 \right ) + b_2 \right )  - X\right \|^2_2 \\
\ell_{sc} &=& \frac{\beta}{m} \left \| \left ( W_1'X + b_1 \right ) - D\alpha  \right \|^2_2 \\ \varepsilon_{en} &=& \frac{\gamma}{m} \left \| \alpha_1 - \alpha_2 \right \|^2_2 
\\
\Omega &=& \lambda \left ( W_1^2 + W_2^2 \right )+ \rho\left \| D \right \|^2_2 +  \frac{\eta}{m}\sqrt{\alpha^2 + \delta } \\\nonumber
\end{eqnarray}

where $\ell_{ae}$ is the auto-encoder loss function, $\ell_{sc}$ is the loss due to the sparse encoding, $\varepsilon_{en}$ is the error of the encoded values and $\Omega$ is the regularization term. $\Omega$ is essential to avoid learning trivial values and to enforce sparsity.


Here, $X = [X^1 \quad X^2]\in\mathbb{R}^{3\times 2m}$,  
where $X^1 = [x^1_1,x^1_2,...,x^1_m]\in\mathbb{R}^{3\times m} $ and $X^2 = [x^2_1,x^2_2,...,x^2_m]\in\mathbb{R}^{3\times m}$ are the RGB values of $m$ randomly sampled pixels from patches extracted from view A images and the corresponding pixels from the patches extracted from view B images respectively. 
$W_1 \in \mathbb{R}^{3\times h}$ is the linear transformation matrix that transforms each of the pixels into a higher dimensional space and $b_1\in \mathbb{R}^{1\times h}$ is the bias term. Similarly, $W_2 \in \mathbb{R}^{h\times 3}$ is the transformation of the higher dimensional space into the original $3$ dimensional space and $b_2\in \mathbb{R}^{1\times 3}$ is the bias term. As mentioned in section \ref{subsec:cc}, previous works have proven that, a linear transformation is sufficient to transform images under an unknown illuminant to images under the canonical illuminant. We borrow this intuition into our work and therefore we use a linear auto-encoder. $D \in \mathbb{R}^{h\times d}$ are the basis vectors(Dictionary or Codebook) to encode each of the transformed pixel values in the higher dimensional space where $d$ is the number of such learned dictionaries. The encoding, $\alpha  = [\alpha_1 \quad \alpha_2]\in\mathbb{R}^{d\times 2m} $ is sparse and is represented by $\alpha_1$ and $\alpha_2$ for $X^1$ and $X^2$ respectively. 

The optimization is done alternatively between $\alpha$, $D$ and $ (W_1,b_1,W_2,b_2)$. We use L-BFGS gradient based optimization procedure to update these values. The gradients with respect to each of the terms are given in Appendix A. Initially, $\alpha_1$ and $\alpha_2$ are updated based on their gradients while keeping $D$ and $ (W_1,b_1,W_2,b_2)$ fixed. Then $D$ is updated keeping $\alpha_1$, $\alpha_2$ and $(W_1,b_1,W_2,b_2)$ fixed. Finally, keeping $\alpha_1$, $\alpha_2$ and $D$ fixed, $(W_1,b_1,W_2,b_2)$ are updated together. For a simple and straightforward gradient based optimization, we replace $\left | \alpha \right |_1$, the L1 norm with an approximation that can smooth it at the origin, $\sqrt{\alpha^2 + \delta }$. $\delta$ is infinitesimally small ($1 \times 10^-4$).

Theoretically, the gradient based optimization is simple for the above objective functions. But for faster convergence and a good optima, it requires a bit of finesse. With that in consideration, practically, good initializations are required for $W_1$, $W_2$ and $D$. We give the objective function for the initializations in Appendix B. Initialization helps to achieve faster convergence as well as a better local optima. Joint learning is done until convergence. During testing, for each pixel in an image, we first use $W_1$ and $b_1$ to transform it into a higher dimensional space and use the learned dictionary $D$ to compute the sparse codes. Therefore, the dimensionality of the features for each of the input image will be $ M \times N \times d$ where $ M \times N \times 3$ is the input image dimensions. 
\figurename{\ref{fig_imgc}} and \figurename{\ref{fig_imgf}} show the representation obtained by us using the joint learning framework. It can be seen that the error of encoding for the pixel sampled from the two patches is much less than the histogram representation of the patch.


\subsection{Multi-layer Framework}

The same formulation can be extended to a multilayer framework so that the representations are close to each other for the patches sampled from the images. To achieve this objective, we encode all the pixels in the sampled patches using the method mentioned in the above section and adopt a max pooling scheme over the $2\times 2$ regions to get the representation of a patch. The {\it max} pooling strategy makes the representation slightly translation invariant. We further adopt the same formulation with the linear transformation and dictionary learning and encode the observations over a patch and enforce the equality constraint over the corresponding samples from the image pairs. Once the representation is obtained in the second layer, max pooling is done over $4\times 4$ regions. For simplicity, we keep the dimensions $h$ and $d$ same for the second layer. We observed that further increasing the number of layers did not give any significant advantage over the performance. Also considering the complexity of the approach, we use only two layers for our model. 
\subsection{Parameters}

The formulation contains several parameters such as the weights of the cost function terms, the dimensions of the linearly transformed space and the number of dictionary atoms. All the parameters were empirically determined by cross-validation. The obtained parameters are as follows. $h=60$, $d=250$, $\beta=1$, $\gamma=0.1$, $\lambda=3\times 10^-3$, $\rho=0.01$ and $\eta=0.01$. 
%
%

\section{Experiments}
\label{sec:exp}
We validate our algorithm on publicly available datasets such as ViPER\cite{viper}, Person Re-ID 2011\cite{hirzer11a} and CAVIAR4REID\cite{custompict}. The characteristics of these three datasets are ideal for the evaluation of the proposed Jointly Learned Color Features(JLCF) since the images were captured from two cameras under varying environments such as indoor and outdoor, bright and dark illumination and different view angles. Below, we list the baseline approaches we compare with.

\begin{enumerate}
\item {\bf Hist}: We compare our approach with the 3D histogram generated in RGB, HSV and YUV spaces. The histograms are computed from the images without any pre-processing for illumination changes. We use 6 bins for each of the channels so that the representation is 216 dimensional which is close to the 250 dimensional space proposed in this work. Image is divided into $8 \times 8$ blocks with a stride of 4 and for each of those blocks, we compute the histogram. We refer to the histogram representations as RGBHist, HSVHist and YUVHist.

\item {\bf cHist}: cHist corresponds to the histogram of the images in the weakly corrected(L2 norm based correction) space. As mentioned for {\bf Hist}, the number of bins for each of the channels and the size of the image blocks are kept same. The representation based on the corrected color space is referred to as cRGBHist, cHSVHist and cYUVHist.

\item {\bf $rg$Hist}: $rg$Hist corresponds to the histogram in the $rg$ space for the corrected images. $rg$ color channels are one of the first photometric invariant color channels proposed. The $rg$ space corresponds to 

\begin{equation}
\label{eqn_rgspace}
r = \frac{R}{R+G+B}, \qquad g = \frac{G}{R+G+B}
\end{equation}
The image is divided into blocks of $8\times 8$ and for each of the blocks, we compute histogram of 16 bins for each channel. The final representation for each block will be 256 dimensional.
\item {\bf Opponent}: The opponent color space is invariant to specularity. It can be computed by 
\begin{equation}
\label{eqn_opponent}
\begin{aligned}[left]
\begin{bmatrix}
O^1\\ 
O^2\\ 
O^3
\end{bmatrix}=\begin{bmatrix}
\frac{1}{\sqrt{2}} &\frac{-1}{\sqrt{2}}  &0 \\ 
\frac{1}{\sqrt{6}} &\frac{1}{\sqrt{6}}  &\frac{-2}{\sqrt{6}} \\ 
\frac{1}{\sqrt{3}} &\frac{1}{\sqrt{3}}  &\frac{1}{\sqrt{3}} 
\end{bmatrix} \begin{bmatrix}
R\\ 
G\\ 
B
\end{bmatrix}
\end{aligned} 
\end{equation}
Image is divided into $8\times 8$ blocks and the final histogram representation is computed as mentioned in \cite{evaluatinggevers2010}.
\item {\bf C}: The C color space adds photometric invariant with respect to shadow shading to the opponent color space. It is computed by normalizing Opponent descriptor by the intensity. 

\begin{equation}
C = \begin{bmatrix}
\frac{O^1}{O^3} &\frac{O^2}{O^3}  &O^3 
\end{bmatrix}^T
\end{equation}

For $8\times 8$ blocks, the histogram is computed as mentioned above for the opponent color space.
\item {\bf Independent Learning}: The learning strategy will be optimizing the auto-encoder transformation first and then obtaining the sparse codes of the transformed pixel values without joint learning. 
The objective given in appendix A for initialization is used to find the optimal auto-encoder transformation. After obtaining the transformation, dictionary is learned using $\ell_{sc}$ in equation \ref{eqn_test}.

\item {\bf JLCF without Color Constancy(JLCF WCC)}: To show that the learning based on color constancy is important, we develop the representation without the color constancy term, i.e., excluding the $\left \| \alpha_1 - \alpha_2 \right \|^2_2$ term in equation \ref{eqn_test}.

\end{enumerate}

The final representation of an image in each of the color space is obtained by concatenating the histogram of each blocks in the image. Keeping the settings same for all of the baselines, we use LADF\cite{ZhenliShiyu_CVPR2013} metric learning framework for all the comparisons. For baseline 6, 7 and the proposed JLCF, the matching score of first and second layer is combined to form the final score.

The results are reported based on the Cumulative Matching Characteristics(CMC)\cite{cmcmoon2001computational}. 
Each of the datasets, experimental settings and their evaluations are given in detail in the following subsections.
Since the color features are complementary to other types of learned low-level and high-level features, we also perform experiments by combining with them to compare with state-of-the-art results. The following features are combined with the proposed color features.

\begin{enumerate}
\item {\bf AE}: Single layer auto-encoder features are learned at a patch of size $8\times 8\times 3$. 400 filters are learned and the filter response of patches for an image are pooled over $8\times 8$ regions. Vectorizing this representation gives the final feature for a single image. The features learned from a single layer auto-encoder are gabor-like edges.
\item {\bf CNN}: The imagenet pre-trained model of Caffe\cite{Jia13caffe} which follows the architecture in \cite{krizhevsky2012imagenet} is used to obtain high-level features. The dimensionality of the obtained feature is $4096$.
\end{enumerate}

\begin{table}[!t]

	\renewcommand{\arraystretch}{1.3}
	
	\caption{Performance Comparison of different baselines and photometric invariant color spaces on the Standard Evaluation Split of the ViPER, Person Re-ID 2011 and CAVIAR4REID Datasets. Proposed Jointly Learned Color Features(JLCF) outperform all the baselines.}
	\label{table_Base}
	\centering
	\subfloat[ViPER]{
	\begin{tabular}{|c|c|c|c|c|c|}
		\hline
		\bfseries Method & \bfseries Rank 1 & \bfseries Rank 5 & \bfseries Rank 10 & \bfseries Rank 15
		\\\hline\hline
		RGBHist  & 7.59 \% & 27.53 \% & 43.35 \% & 54.74 \% \\
		\hline
		HSVHist  & 11.07 \% & 33.23 \% & 48.42 \% &  60.44 \%\\
		\hline
		YUVHist  & 1.90 \% & 10.44 \% & 18.35 \% & 24.05 \%\\
		\hline
		cRGBHist & 7.59 \% & 27.53 \% & 43.67 \% & 54.74 \%\\
		\hline
		cHSVHist & 12.34 \% & 38.29 \% & 52.53 \% & 65.19 \%\\
		\hline
		cYUVHist & 3.48 \% & 13.60 \% & 22.46 \% & 28.16 \%\\
		\hline
		rgHist  & 1.90 \% & 7.59 \% & 16.46 \% & 22.78 \% \\
				\hline
				C Color Space  & 7.59 \% & 26.27 \% & 40.50 \% &  48.73 \%\\
				\hline
				Opponent  & 7.91 \% & 25.31 \% & 38.29 \% & 43.67 \%\\
				\hline
				{\bf \begin{tabular}{@{}c@{}}Independent \\ Learning\end{tabular}} & 20.25 \%  & 47.78 \% & 64.24 \% & 76.89 \%\\
				\hline
				{\bf JLCF WCC} & 19.94 \% & 48.73 \% & 65.50 \% & 75.00 \%\\
				\hline
		{\bf Proposed JLCF} & {\bf 26.27 \%} & {\bf 51.90 \%} & {\bf 67.09 \%} & {\bf 78.17 \%}\\
		\hline
	\end{tabular}	}

	\subfloat[Person Re-ID 2011]{
	\begin{tabular}{|c|c|c|c|c|c|}
			\hline
			\bfseries Method & \bfseries Rank 1 & \bfseries Rank 10 & \bfseries Rank 20 & \bfseries Rank 50
			\\\hline\hline
			RGBHist  & 1 \% & 14 \% & 25 \% & 41 \%\\
			\hline
			HSVHist  & 0 \% & 7 \% & 14 \% & 17 \%\\
			\hline
			YUVHist  & 1 \% & 6 \% & 12 \% & 27 \%\\
			\hline
			cRGBHist & 1 \% & 13 \% & 25 \% & 42 \%\\
			\hline
			cHSVHist & 0 \% & 11 \% & 22 \% & 31 \%\\
			\hline
			cYUVHist & 3 \% & 9 \% & 14 \% & 33 \%\\
			\hline
			rgHist  & 0 \% & 2 \% & 6 \% & 21 \% \\
					\hline
					C Color Space   & 3 \% & 12 \% & 20 \% & 33 \% \\
					\hline
					Opponent  & 3 \% & 21 \% & 30 \% & 42 \% \\
					\hline
					{\bf \begin{tabular}{@{}c@{}}Independent \\ Learning\end{tabular}} & 6 \%  & 32 \% & 41 \% & 62 \%\\
					\hline
					{\bf JLCF WCC} & 9 \% & 36 \% & 47 \% & 60 \%\\
					\hline
			{\bf Proposed JLCF} & {\bf 14 \%} & {\bf 45 \%} & {\bf 55 \%} & {\bf 72 \%}\\
			\hline
		\end{tabular}
	}

	\subfloat[CAVIAR4REID]{
	\begin{tabular}{|c|c|c|c|c|}
			\hline
			\bfseries Method & \bfseries Rank 1 & \bfseries Rank 5 & \bfseries Rank 10 & \bfseries Rank 20
			\\\hline\hline
			RGBHist  & 19.79 \% & 65.47 \% & 81.89 \% & 98.94 \% \\
			\hline
			HSVHist  & 18.73 \% & 56.42 \% & 76.00 \% &  93.68 \% \\
			\hline
			YUVHist  & 17.05 \% & 51.36 \% & 68.84 \% & 94.52 \% \\
			\hline
			cRGBHist & 24.21 \% & 68.42 \% & 84.84 \% & 97.47 \% \\
			\hline
			cHSVHist & 20.21 \% & 59.36 \% & 77.47 \% & 94.94 \%\\
			\hline
			cYUVHist & 22.73 \% & 53.26 \% & 72.42 \% & 96.84 \% \\
			\hline
			rgHist  & 12.84 \% & 43.36 \%  & 68.84 \% & 92.63 \%\\
					\hline
					C Color Space  & 22.52 \%  & 56.63 \% &  72.21 \% & 94.73 \%\\
					\hline
					Opponent  & 25.05 \%  & 58.10 \% & 71.78 \% & 90.52 \%\\
					\hline
					{\bf \begin{tabular}{@{}c@{}}Independent \\ Learning\end{tabular}} & 25.47 \%  & 66.52 \% & 86.526 \% & 99.36 \%\\
					\hline
					{\bf JLCF WCC} & 26.68 \%  & 65.47 \% & 82.73 \% & 98.31 \%\\
					\hline
			{\bf Proposed JLCF} & {\bf 32.63 \%} & {\bf 67.15 \%} & {\bf 87.57 \%} & {\bf 99.36 \%} \\
			\hline
		\end{tabular}
	}	
\end{table}

\subsection{ViPER Dataset}

ViPER\cite{viper} is the most popular and challenging dataset to evaluate Person Re-Identification. The dataset contains 632 pedestrians from arbitrary viewpoints under varying illumination conditions and have relatively low resolution. The images are normalized to a size of $128\times 48$. We use the same settings as mentioned in \cite{zhao2013unsupervised,ZhenliShiyu_CVPR2013} for the evaluation.

Table \ref{table_Base} shows the performance comparison of our approach with different baseline methods and photometric invariant color spaces. Experimental results suggest that the encoding based on the joint learning helps to a very large extend in achieving good performance. Table \ref{table_Base} also shows our comparison with the independent learning strategy as well as the joint learning framework without the color constancy term, $\varepsilon_{en}$ in equation \ref{eqn_test}.

An evaluation of the state of the art algorithms is given in table \ref{table_viperState}. The \cite{ZhenliShiyu_CVPR2013}, \cite{zhang2014novel} and \cite{zhang2014structured} out performs our methods at different ranks. We observed that this is due to the lack of an optimal score combining mechanism for color and texture features. \cite{Zhao_2014_CVPR} achieved state-of-the-art results by combining with LADF. However, using color as a single cue, we achieve comparable results with several state-of-the-art methods based on multiple handcrafted features.
\subsection{Person Re-ID 2011}
The Person Re-ID 2011 dataset\cite{hirzer11a} consists of images extracted from multiple person trajectories recorded from two different, static surveillance cameras. Images from
these cameras contain a sheer difference in illumination, background, camera characteristics and a significant view point change. Multiple images per person are available in each camera view. There are 385 person trajectories from one view, 749 from the other and 200 people appear in both views. For more details regarding the dataset, we refer the reader to \cite{hirzer11a} and \cite{hirzer2012relaxed}.


In our experiments we do a multi-shot re-id with the same settings as mentioned in \cite{hirzer2012relaxed} and compare our results with the baselines as well as the state-of-the-art results. Table \ref{table_Base} shows the comparison of our approach with the baselines and photometric invariant color spaces. Table \ref{table_viperState} shows the comparison of our approach with state-of-the-art results in Person Re-ID 2011 dataset. As shown in the results, our method clearly outperforms the baselines, different photometric invariant color spaces and when combined with AE and CNN features, it outperforms the state-of-the-art results.

\subsection{CAVIAR4REID}
CAVIAR4REID\cite{custompict} is a person re-id evaluation dataset which was extracted from the well known caviar dataset for evaluation of people tracking and detection algorithms. It is a relatively smaller dataset which contains a total of 72 pedestrians(50 of them in both camera views and the remaining 22 with one camera only) taken from a shopping mall in Lisbon. Re-identification in this dataset is challenging due to a large variation in the resolution, illumination, occlusion and pose changes. 

The experimental settings are kept the same as in \cite{kviatkovsky2013color}. Table \ref{table_Base} shows that the illumination invariant signatures are performing significantly better than the baseline color features and different photometric invariant color spaces. We also compare with other standard approaches by combining JLCF with AE and CNN features and the results are reported in table \ref{table_viperState}. Similar to the ViPER dataset, we observed that lack of an optimal score combining mechanism affects the performance at rank 1. However, it should be noted that we achieve the best results at higher ranks.

\begin{table}[!t]
	\renewcommand{\arraystretch}{1.3}
	\caption{Performance Comparison of different state-of-the-art results on the Standard Evaluation Split of the ViPER, Person Re-ID 2011 and CAVIAR4RED Datasets. Proposed Jointly Learned Color Features(JLCF) combined with AE and CNN can achieve promising results in the following datasets.}
	\label{table_viperState}
	\centering
	\subfloat[ViPER]{
	\begin{tabular}{|c|c|c|c|c|}
		\hline
		\bfseries Method & \bfseries Rank 1 & \bfseries Rank 5 & \bfseries Rank 10 & \bfseries Rank 15 
		\\\hline\hline
		SDALF\cite{sdalf}  & 19.87 \% & 38.89 \% & 49.37 \% & 58.46 \% \\
		\hline
		CPS\cite{custompict}  & 21.84 \% & 44.00 \% & 57.21 \% &  65.18 \% \\
		\hline
		eBiCov\cite{bicovma2012}  & 20.66 \% & 42.00 \% & 56.18 \% & 63.11 \% \\
		\hline
		ELF\cite{viewpointinvariant}  & 12 \% & 41.50 \% & 59.50 \% & 68.00 \% \\
		\hline		
		SalMatch\cite{zhao2013person} & 30.15 \% & 52.31 \% & 65.53 \% & 73.41 \%\\
		\hline
		PatMatch\cite{zhao2013unsupervised} & 26.90 \% & 47.46 \% & 62.34 \% & 73.41 \%\\
		\hline	
		\begin{tabular}{@{}c@{}} Mid-level \\Features\cite{Zhao_2014_CVPR}\end{tabular} & 29.11 \% & 52.34 \% & 65.95 \% & 73.92 \%\\
		\hline
		LADF\cite{ZhenliShiyu_CVPR2013} & 29.43 \% & 63.29 \% &  76.27 \% & 83.23 \% \\
		\hline
		VWCM\cite{zhang2014novel} & 30.70 \% & 62.97 \% &  75.95 \% & 81.01 \% \\
		\hline
		PRSP\cite{zhang2014structured} & 38.92 \% & 67.41 \% & 80.38 \% &  84.81 \% \\
		\hline
		{\bf Proposed JLCF} & 26.27 \% &  51.90 \% &  67.09 \% &  78.17 \%\\
		\hline	
		{\bf\begin{tabular}{@{}c@{}} Proposed JLCF + \\ AE + CNN\end{tabular}} &  32.28 \% & 56.02 \% & 70.26 \% & 79.75 \% \\
		\hline
	\end{tabular}
	} \\
	\subfloat[Person Re-ID 2011]{
	\begin{tabular}{|c|c|c|c|c|}
		\hline
		\bfseries Method & \bfseries Rank 1 & \bfseries Rank 10 & \bfseries Rank 20 & \bfseries Rank 50 
		\\\hline\hline
		Descr. Model\cite{hirzer11a}  & 4 \% & 24 \% & 37 \% & 56 \% \\
		\hline
		RPML\cite{hirzer2012relaxed}  & 15 \% & 42 \% & 54 \% &  70 \%\\
		\hline
		{\bf Proposed JLCF} & 14 \% &  45 \% &  55 \% &  72 \%\\
		\hline	
		{\bf\begin{tabular}{@{}c@{}} Proposed JLCF + \\ AE + CNN\end{tabular}} & {\bf 19 \%} & {\bf 47 \%} & {\bf 57 \%} & {\bf 75 \%}\\
		\hline
	\end{tabular}	
	}
	\\ 
	
	\subfloat[CAVIAR4REID]{
	\begin{tabular}{|c|c|c|c|c|c|}
		\hline
		\bfseries Method & \bfseries Rank 1 & \bfseries Rank 5 & \bfseries Rank 10 & \bfseries Rank 20
		\\\hline\hline
		HPE\cite{Bazzani2012898}  & 9.7 \%  & 33.2 \% & 55.6 \% & 76.3 \%\\
		\hline
		LF\cite{pedagadilfda}  & 36.1 \%  & 51.2 \% &  88.6 \% & 97.5 \%\\
		\hline
		LADF\cite{ZhenliShiyu_CVPR2013} & 29.64 \% & 62.01 \% & 78.52 \% & 94.23 \%\\
		\hline	
		SSCDL\cite{Liu_2014_CVPR} & {\bf 49.1}\% & 80.2 \% & 93.5 \% & 97.9 \%\\
		\hline	
		{\bf Proposed JLCF} &  32.63 \% &  67.15 \% &  87.57 \% &  99.36 \% \\
		\hline
		{\bf\begin{tabular}{@{}c@{}} Proposed JLCF + \\ AE + CNN\end{tabular}} &  45.89 \% & {\bf 80.84 \%} & {\bf 94.10 \%} & {\bf 100.00 \%}\\
		\hline                                                                                        
	\end{tabular}	
	}	
\end{table}

\section{Analysis}
In this section, we give the analysis of our approach and compare it with the baseline methods for color features.
\subsection{JLCF vs Designed color spaces}
Color histograms are representations which capture the color distribution. However, the difference in illumination can cause a significant change in the appearance which is caused by the variation in the RGB pixel values. Therefore, without the illumination correction, the histogram will not be a robust representation for color images. Using L2-norm, we do a correction for each of the images and then obtain the histogram in such a corrected space, the {\bf cHist}. Our approach uses illumination correction and find an optimal transformation to encode the pixel values in such a way that pixels corresponding to similar colors are close enough. The other photometric invariant color spaces can be considered as handcrafted features addressing specific cues as mentioned in section \ref{sec:exp}. Table \ref{table_Base} shows the comparisons of our method with the baseline approaches and it can be seen that JLCF clearly outperforms all of them. This is due to the fact that the L2 norm based correction which is inspired from the diagonal model is a weak illumination correction due to the strong assumptions. The comparison also shows that data driven learning approach is better than handcrafted color features.
\subsection{JLCF vs Independent Learning}
As mentioned in section \ref{sec:motiv}, to be consistent with each other, the linear auto-encoder transformation and dictionary for sparse coding must be learned jointly. As it can be seen from the results in table \ref{table_Base}, the joint learning improves the performance significantly for all the datasets. This is due to the fact that, for the encoding of each pixel, an optimal dictionary which can give same representation for pixels of same color has to be learned together with the linear transformation.
\subsection{Proposed JLCF vs JLCF without Color Constancy}
The objective of our approach is that for similar colors, the encoding must be same. Since the pixel values are corrupted by varying lighting conditions, this cannot be achieved by merely taking a histogram in pre-defined color spaces or sparse coding unless the pixel values are {\it close enough}. With this objective, the formulation in equation \ref{eqn_test} encodes the color information in such a way that for same colors, the encoding error is minimized. We conduct experiments with and without the color constancy term ($\varepsilon_{en}$) and report the results in table \ref{table_Base}. It can be seen that the color constancy based encoding improves the performance significantly. This is due to the fact that, without the color constancy term, the objective function merely encodes the pixels in a new space without any correction for the varying lighting conditions. But it should also be noted that the sparse encoding of linearly transformed pixel values results in a much better representation than histograms.

\section{Conclusion}
In this paper, we propose a novel data driven framework for learning color features to handle illumination and other lighting condition changes across two camera views. In contrast to the previous works based on auto-encoders and sparse coding, we combine them to learn a robust encoding jointly by forcing similar colors to be close to each other. We have also evaluated several baseline methods for achieving color constancy and have shown superior performance over all of them. By combining with other types of learned low-level and high-level features, we achieve promising results in several benchmark datasets.
\ifCLASSOPTIONcaptionsoff
  \newpage
\fi



\bibliographystyle{IEEEtran}
\bibliography{IEEEabrv,references}

\begin{thebibliography}{10}
\providecommand{\url}[1]{#1}
\csname url@samestyle\endcsname
\providecommand{\newblock}{\relax}
\providecommand{\bibinfo}[2]{#2}
\providecommand{\BIBentrySTDinterwordspacing}{\spaceskip=0pt\relax}
\providecommand{\BIBentryALTinterwordstretchfactor}{4}
\providecommand{\BIBentryALTinterwordspacing}{\spaceskip=\fontdimen2\font plus
\BIBentryALTinterwordstretchfactor\fontdimen3\font minus
  \fontdimen4\font\relax}
\providecommand{\BIBforeignlanguage}[2]{{%
\expandafter\ifx\csname l@#1\endcsname\relax
\typeout{** WARNING: IEEEtran.bst: No hyphenation pattern has been}%
\typeout{** loaded for the language `#1'. Using the pattern for}%
\typeout{** the default language instead.}%
\else
\language=\csname l@#1\endcsname
\fi
#2}}
\providecommand{\BIBdecl}{\relax}
\BIBdecl

\bibitem{kviatkovsky2013color}
I.~Kviatkovsky, A.~Adam, and E.~Rivlin, ``Color invariants for person
  reidentification,'' \emph{Pattern Analysis and Machine Intelligence, IEEE
  Transactions on}, vol.~35, no.~7, pp. 1622--1634, July 2013.

\bibitem{custompict}
D.~S. Cheng, M.~Cristani, M.~Stoppa, L.~Bazzani, and V.~Murino, ``Custom
  pictorial structures for re-identification,'' in \emph{Proceedings of the
  British Machine Vision Conference}.\hskip 1em plus 0.5em minus 0.4em\relax
  BMVA Press, 2011, pp. 68.1--68.11.

\bibitem{bicovma2012}
B.~Ma, Y.~Su, and F.~Jurie, ``\BIBforeignlanguage{English}{{BiCov: a novel
  image representation for person re-identification and face verification}},''
  in \emph{\BIBforeignlanguage{English}{{British Machive Vision Conference}}},
  Guildford, United Kingdom, 2012, p. 11 pages.

\bibitem{viewpointinvariant}
D.~Gray and H.~Tao, ``Viewpoint invariant pedestrian recognition with an
  ensemble of localized features,'' in \emph{Computer Vision – ECCV 2008},
  ser. Lecture Notes in Computer Science, D.~Forsyth, P.~Torr, and
  A.~Zisserman, Eds.\hskip 1em plus 0.5em minus 0.4em\relax Springer Berlin
  Heidelberg, 2008, vol. 5302, pp. 262--275.

\bibitem{viper}
D.~Gray, S.~Brennan, and H.~Tao, ``Evaluating appearance models for
  recognition, reacquisition, and tracking,'' in \emph{10th IEEE International
  Workshop on Performance Evaluation of Tracking and Surveillance (PETS)},
  09/2007 2007.

\bibitem{hirzer11a}
M.~Hirzer, C.~Beleznai, P.~M. Roth, and H.~Bischof, ``Person re-identification
  by descriptive and discriminative classification,'' in \emph{Proc.
  Scandinavian Conf. on Image Analysis}, 2011, the original publication is
  available at www.springerlink.com.

\bibitem{fastimplofccmorel2009}
J.-M. Morel, A.~B. Petro, and C.~Sbert, ``Fast implementation of color
  constancy algorithms,'' \emph{Proceedings of the SPIE}, vol. 7241, 2009.

\bibitem{retinexland71}
E.~H. LAND and J.~J. McCANN, ``Lightness and retinex theory,'' \emph{J. Opt.
  Soc. Am.}, vol.~61, no.~1, pp. 1--11, Jan 1971.

\bibitem{forsythcolorconst}
D.~Forsyth, ``\BIBforeignlanguage{English}{A novel algorithm for color
  constancy},'' \emph{\BIBforeignlanguage{English}{International Journal of
  Computer Vision}}, vol.~5, no.~1, pp. 5--35, 1990.

\bibitem{ccsurvey2011}
A.~Gijsenij, T.~Gevers, and J.~van~de Weijer, ``Computational color constancy:
  Survey and experiments,'' \emph{Image Processing, IEEE Transactions on},
  vol.~20, no.~9, pp. 2475--2489, Sept 2011.

\bibitem{ebner2007color}
M.~Ebner, \emph{Color constancy}.\hskip 1em plus 0.5em minus 0.4em\relax John
  Wiley \& Sons, 2007, vol.~6.

\bibitem{hordley2006scene}
S.~D. Hordley, ``Scene illuminant estimation: past, present, and future,''
  \emph{Color Research \& Application}, vol.~31, no.~4, pp. 303--314, 2006.

\bibitem{swainballardcolorhistogram}
M.~Swain and D.~Ballard, ``Indexing via color histograms,'' in \emph{Computer
  Vision, 1990. Proceedings, Third International Conference on}, Dec 1990, pp.
  390--393.

\bibitem{Geverscolorbasedobj}
T.~Gevers and A.~W. Smeulders, ``Color-based object recognition,''
  \emph{Pattern Recognition}, vol.~32, no.~3, pp. 453 -- 464, 1999.

\bibitem{berwicklee}
D.~Berwick and S.~W. Lee, ``A chromaticity space for specularity, illumination
  color- and illumination pose-invariant 3-d object recognition,'' in
  \emph{Computer Vision, 1998. Sixth International Conference on}, Jan 1998,
  pp. 165--170.

\bibitem{healey1994global}
G.~Healey and D.~Slater, ``Global color constancy: recognition of objects by
  use of illumination-invariant properties of color distributions,'' \emph{JOSA
  A}, vol.~11, no.~11, pp. 3003--3010, 1994.

\bibitem{Finlayson94}
G.~D. Finlayson, M.~S. Drew, and B.~V. Funt, ``Spectral sharpening: sensor
  transformations for improved color constancy,'' \emph{J. Opt. Soc. Am. A},
  vol.~11, no.~5, pp. 1553--1563, May 1994.

\bibitem{sdalf}
M.~Farenzena, L.~Bazzani, A.~Perina, V.~Murino, and M.~Cristani, ``Person
  re-identification by symmetry-driven accumulation of local features,'' in
  \emph{Proceedings of the 2010 IEEE Computer Society Conference on Computer
  Vision and Pattern Recognition (CVPR 2010)}.\hskip 1em plus 0.5em minus
  0.4em\relax San Francisco, CA, USA: IEEE Computer Society, 2010.

\bibitem{zhao2013unsupervised}
R.~Zhao, W.~Ouyang, and X.~Wang, ``Unsupervised salience learning for person
  re-identification,'' in \emph{Computer Vision and Pattern Recognition (CVPR),
  2013 IEEE Conference on}.\hskip 1em plus 0.5em minus 0.4em\relax IEEE, 2013,
  pp. 3586--3593.

\bibitem{btfporikli2003}
F.~Porikli, ``Inter-camera color calibration by correlation model function,''
  in \emph{Image Processing, 2003. ICIP 2003. Proceedings. 2003 International
  Conference on}, vol.~2, Sept 2003, pp. II--133--6 vol.3.

\bibitem{btfjaved2005}
O.~Javed, K.~Shafique, and M.~Shah, ``Appearance modeling for tracking in
  multiple non-overlapping cameras,'' in \emph{Computer Vision and Pattern
  Recognition, 2005. CVPR 2005. IEEE Computer Society Conference on}, vol.~2,
  June 2005, pp. 26--33 vol. 2.

\bibitem{vincent2008extracting}
P.~Vincent, H.~Larochelle, Y.~Bengio, and P.-A. Manzagol, ``Extracting and
  composing robust features with denoising autoencoders,'' in \emph{Proceedings
  of the 25th international conference on Machine learning}.\hskip 1em plus
  0.5em minus 0.4em\relax ACM, 2008, pp. 1096--1103.

\bibitem{yang2009linear}
J.~Yang, K.~Yu, Y.~Gong, and T.~Huang, ``Linear spatial pyramid matching using
  sparse coding for image classification,'' in \emph{Computer Vision and
  Pattern Recognition, 2009. CVPR 2009. IEEE Conference on}.\hskip 1em plus
  0.5em minus 0.4em\relax IEEE, 2009, pp. 1794--1801.

\bibitem{hinton2006fast}
G.~Hinton, S.~Osindero, and Y.-W. Teh, ``A fast learning algorithm for deep
  belief nets,'' \emph{Neural computation}, vol.~18, no.~7, pp. 1527--1554,
  2006.

\bibitem{hinton2006reducing}
G.~E. Hinton and R.~R. Salakhutdinov, ``Reducing the dimensionality of data
  with neural networks,'' \emph{Science}, vol. 313, no. 5786, pp. 504--507,
  2006.

\bibitem{lee2008sparse}
H.~Lee, C.~Ekanadham, and A.~Y. Ng, ``Sparse deep belief net model for visual
  area v2,'' in \emph{Advances in neural information processing systems}, 2008,
  pp. 873--880.

\bibitem{bengio2007greedy}
Y.~Bengio, P.~Lamblin, D.~Popovici, and H.~Larochelle, ``Greedy layer-wise
  training of deep networks,'' \emph{Advances in neural information processing
  systems}, vol.~19, p. 153, 2007.

\bibitem{bengio2009learning}
Y.~Bengio, ``Learning deep architectures for ai,'' \emph{Foundations and
  trends{\textregistered} in Machine Learning}, vol.~2, no.~1, pp. 1--127,
  2009.

\bibitem{serre2007quantitative}
T.~Serre, G.~Kreiman, M.~Kouh, C.~Cadieu, U.~Knoblich, and T.~Poggio, ``A
  quantitative theory of immediate visual recognition,'' \emph{Progress in
  brain research}, vol. 165, pp. 33--56, 2007.

\bibitem{yu2009nonlinear}
K.~Yu, T.~Zhang, and Y.~Gong, ``Nonlinear learning using local coordinate
  coding,'' in \emph{Advances in neural information processing systems}, 2009,
  pp. 2223--2231.

\bibitem{weakinvariancedrew1998}
M.~Drew, J.~Wei, and Z.-N. Li, ``Illumination-invariant color object
  recognition via compressed chromaticity histograms of
  color-channel-normalized images,'' in \emph{Computer Vision, 1998. Sixth
  International Conference on}, Jan 1998, pp. 533--540.

\bibitem{evaluatinggevers2010}
K.~E.~A. Van~de Sande, T.~Gevers, and C.~G.~M. Snoek, ``Evaluating color
  descriptors for object and scene recognition,'' \emph{Pattern Analysis and
  Machine Intelligence, IEEE Transactions on}, vol.~32, no.~9, pp. 1582--1596,
  Sept 2010.

\bibitem{ZhenliShiyu_CVPR2013}
Z.~Li, S.~Chang, F.~Liang, T.~S. Huang, L.~Cao, and J.~R. Smith, ``Learning
  locally-adaptive decision functions for person verification,'' in \emph{IEEE
  Conference on Computer Vision and Pattern Recognition (CVPR)}, June 2013.

\bibitem{cmcmoon2001computational}
H.~Moon and P.~J. Phillips, ``Computational and performance aspects of
  pca-based face-recognition algorithms,'' \emph{Perception-London}, vol.~30,
  no.~3, pp. 303--322, 2001.

\bibitem{Jia13caffe}
Y.~Jia, ``{Caffe}: An open source convolutional architecture for fast feature
  embedding,'' {http://caffe.berkeleyvision.org/}, 2013.

\bibitem{krizhevsky2012imagenet}
A.~Krizhevsky, I.~Sutskever, and G.~E. Hinton, ``Imagenet classification with
  deep convolutional neural networks,'' in \emph{Advances in neural information
  processing systems}, 2012, pp. 1097--1105.

\bibitem{zhang2014novel}
Z.~Zhang, Y.~Chen, and V.~Saligrama, ``A novel visual word co-occurrence model
  for person re-identification,'' in \emph{ECCV Workshop on Visual Surveillance
  and Re-Identification}, 2014.

\bibitem{zhang2014structured}
\BIBentryALTinterwordspacing
Z.~Zhang and V.~Saligrama, ``Person re-identification via structured
  prediction,'' \emph{CoRR}, vol. abs/1406.4444, 2014. [Online]. Available:
  \url{http://arxiv.org/abs/1406.4444}
\BIBentrySTDinterwordspacing

\bibitem{Zhao_2014_CVPR}
R.~Zhao, W.~Ouyang, and X.~Wang, ``Learning mid-level filters for person
  re-identification,'' in \emph{The IEEE Conference on Computer Vision and
  Pattern Recognition (CVPR)}, June 2014.

\bibitem{hirzer2012relaxed}
M.~Hirzer, P.~M. Roth, M.~K{\"o}stinger, and H.~Bischof, ``Relaxed pairwise
  learned metric for person re-identification,'' in \emph{Computer Vision--ECCV
  2012}.\hskip 1em plus 0.5em minus 0.4em\relax Springer, 2012, pp. 780--793.

\bibitem{zhao2013person}
R.~Zhao, W.~Ouyang, and X.~Wang, ``Person re-identification by salience
  matching,'' in \emph{IEEE International Conference on Computer Vision
  (ICCV)}, Sydney, Australia, December 2013.

\bibitem{Bazzani2012898}
L.~Bazzani, M.~Cristani, A.~Perina, and V.~Murino, ``Multiple-shot person
  re-identification by chromatic and epitomic analyses,'' \emph{Pattern
  Recognition Letters}, vol.~33, no.~7, pp. 898 -- 903, 2012, special Issue on
  Awards from \{ICPR\} 2010.

\bibitem{pedagadilfda}
S.~Pedagadi, J.~Orwell, S.~Velastin, and B.~Boghossian, ``Local fisher
  discriminant analysis for pedestrian re-identification,'' \emph{2013 IEEE
  Conference on Computer Vision and Pattern Recognition}, vol.~0, pp.
  3318--3325, 2013.

\bibitem{Liu_2014_CVPR}
X.~Liu, M.~Song, D.~Tao, X.~Zhou, C.~Chen, and J.~Bu, ``Semi-supervised coupled
  dictionary learning for person re-identification,'' in \emph{The IEEE
  Conference on Computer Vision and Pattern Recognition (CVPR)}, June 2014.

\end{thebibliography}
%

%
%

%

\ifCLASSOPTIONtechnote
\else
	\begin{IEEEbiography}{Michael Shell}
	Biography text here.
	\end{IEEEbiography}
	
	\begin{IEEEbiographynophoto}{John Doe}
	Biography text here.
	\end{IEEEbiographynophoto}
	
	
	\begin{IEEEbiographynophoto}{Jane Doe}
	Biography text here.
	\end{IEEEbiographynophoto}

\fi


\end{document}